\documentclass[times,twocolumn,final]{elsarticle}

\usepackage{medima}
\usepackage{framed,multirow}

\usepackage{amsmath}
\usepackage{amssymb}
\usepackage{latexsym}
\usepackage{algorithm}
\usepackage{array}
\usepackage{graphicx}
\usepackage[table,xcdraw]{xcolor}
\usepackage{algpseudocode}

\usepackage{url}
\usepackage{xcolor}

\usepackage{hyperref}

\definecolor{newcolor}{rgb}{.8,.349,.1}

\journal{Medical Image Analysis}

\DeclareMathOperator*{\argmin}{arg\,min}

\newcommand{\mat}[1]{\mathbf{#1}}

\DeclareMathOperator*{\argmax}{arg\,max}
\DeclareMathOperator*{\median}{median}

\begin{document}

\verso{Javad Fotouhi \textit{et~al.}}

\begin{frontmatter}

\title{Exploring Partial Intrinsic and Extrinsic Symmetry in 3D Medical Imaging \tnoteref{tnote1}}

\author[1]{Javad \snm{Fotouhi}\corref{cor1}}
\cortext[cor1]{Corresponding author}
\ead{javad.fotouhi@jhu.edu}
\author[1]{Giacomo \snm{Taylor}\fnref{fn1}}
\author[1]{Mathias \snm{Unberath}}
\author[3]{Alex \snm{Johnson M.D.}}
\author[1]{Sing Chun \snm{Lee}}
\author[3]{Greg \snm{Osgood M.D.}}
\author[2,3]{Mehran \snm{Armand} \corref{cor2} \cortext[cor2]{Joint senior authors in alphabetical order}}
\author[1,4]{Nassir \snm{Navab} \corref{cor2}}

\address[1]{Department of Computer Science, Johns Hopkins University, Baltimore, MD, USA}
\address[2]{Applied Physics Laboratory, Johns Hopkins University, Laurel, MD, USA}
\address[3]{Department of Orthopedic Surgery, Johns Hopkins Hospital, Baltimore, USA}
\address[4]{Laboratory for Computer Aided Medical Procedures, Technical University of Munich, Munich, Germany}

\received{* ** 2020}
\finalform{* ** ****}
\accepted{* ** ****}
\availableonline{* ** ***}
\communicated{****}

\begin{abstract}
We present a novel methodology to detect imperfect bilateral symmetry in CT of human anatomy. In this paper, the structurally symmetric nature of the pelvic bone is explored and is used to provide interventional image augmentation for treatment of unilateral fractures in patients with traumatic injuries. The mathematical basis of our solution is on the incorporation of attributes and characteristics that satisfy the properties of intrinsic and extrinsic symmetry and are robust to outliers. In the first step, feature points that satisfy intrinsic symmetry are automatically detected in the M\"obius space defined on the CT data. These features are then pruned via a two-stage RANSAC to attain correspondences that satisfy also the extrinsic symmetry. Then, a disparity function based on Tukey's biweight robust estimator is introduced and minimized to identify a symmetry plane parametrization that yields maximum contralateral similarity. Finally, a novel regularization term is introduced to enhance similarity between bone density histograms across the partial symmetry plane, relying on the important biological observation that, even if injured, the dislocated bone segments remain within the body. Our extensive evaluations on various cases of common fracture types demonstrate the validity of the novel concepts and the robustness and accuracy of the proposed method.
\end{abstract}


\end{frontmatter}


\section{Introduction}\label{sec:introduction}
Symmetry is an integral property of nature and is ubiquitous in human anatomy and living organisms. For instance, there is considerable amount of structural correlation across the sagittal plane of the human pelvis. Quantitative analysis of healthy pelvis data indicate that $78.9\%$ of the distinguishable anatomical landmarks on the pelvis are symmetric~\cite{boulay2006three}, and the asymmetry in the remaining landmarks are still tolerated for orthopedic surgeries~\cite{shen2013augmented}. Another recent study concluded that the human pelvis exhibits strong bilateral symmetry, and can be used for fracture reconstruction~\cite{ead2020investigation}. 

In the remainder of this section we highlight previous works that aimed at detecting symmetry in shapes and images (Sec.~\ref{subsec:relatedWork}),
present the importance of symmetry in surgical practice (Sec.~\ref{subsec:clinicalMotiv}), 
and propose a novel methodology for exploiting partial symmetry in human pelvis with to incorporate the knowledge from symmetry and augment surgeon's information during fracture care procedures (Sec.~\ref{subsec:proposedSolution}).

\subsection{Related Work}~\label{subsec:relatedWork}

There is a great body of work in the computer vision community that investigated symmetries in 2D images~\cite{liu2010computational}. The knowledge from symmetry has found several applications, namely in depth estimation~\cite{mukherjee1995shape}, detecting camera projections~\cite{gao2017exploiting}, single-view scene reconstruction~\cite{hong2004symmetry}, and image segmentation~\cite{nagar2017symmslic}.

Symmetry recognition has leveraged the success of feature detection methods in computer vision, and used image-based key-points to identify local and global symmetries. The approach by Loy and Eklundh~\cite{loy2006detecting} detects symmetric pair of points and forms local constellations of symmetries that collectively describe a global symmetry. This method searches for symmetries on multiple scales, and all orientations and locations. Bilateral symmetry can also be identified in 2D images under affine and perspective transformations by investigating vanishing points that link quadruplets of feature points in an image~\cite{cornelius2006detecting,cornelius2007efficient}. To increase invariance to local illumination and improve robustness, affine invariant edge-based features were suggested as replacement of intensity-based features to locate planar symmetry, where each edge correspondence casts a vote to find the dominant reflection axis~\cite{wang2015reflection}. The work by Lee and Liu~\cite{lee2011curved} studied glide-reflection, a combination of translation and reflection. Recent works have primarily focused on using convolutional neural networks to predict symmetries~\cite{brachmann2016using,funk2017beyond}. 

Despite the advancements in symmetry detection in 2D images, the immediate translation of such techniques to medical imagery data is yet unclear due to different properties and use-cases. For instance, X-ray transmission imaging, in contrast to reflective imaging, is based on different X-ray attenuation from different tissue. A single pixel in an X-ray image relates to all 3D points along the ray. As a result of attenuation-based physics governing the image formation, depending on the viewing direction, anatomical landmarks may vanish or change appearance. Hence, classic feature detection methods fail to identify key-points~\cite{bier2018x}, and symmetry detection based on feature points can become unreliable. Additionally, symmetry recognition is particularly useful for medical and interventional imaging where the knowledge of symmetry can transfer to 3D, and enable an understanding of the geometry of the contralateral side. This requires a more complex parameterization beyond detecting an in-plane axis of symmetry~\cite{liu2013symmetry, hogeweg2017fast}.

\begin{figure}
	\centering
		\includegraphics[width=\columnwidth]{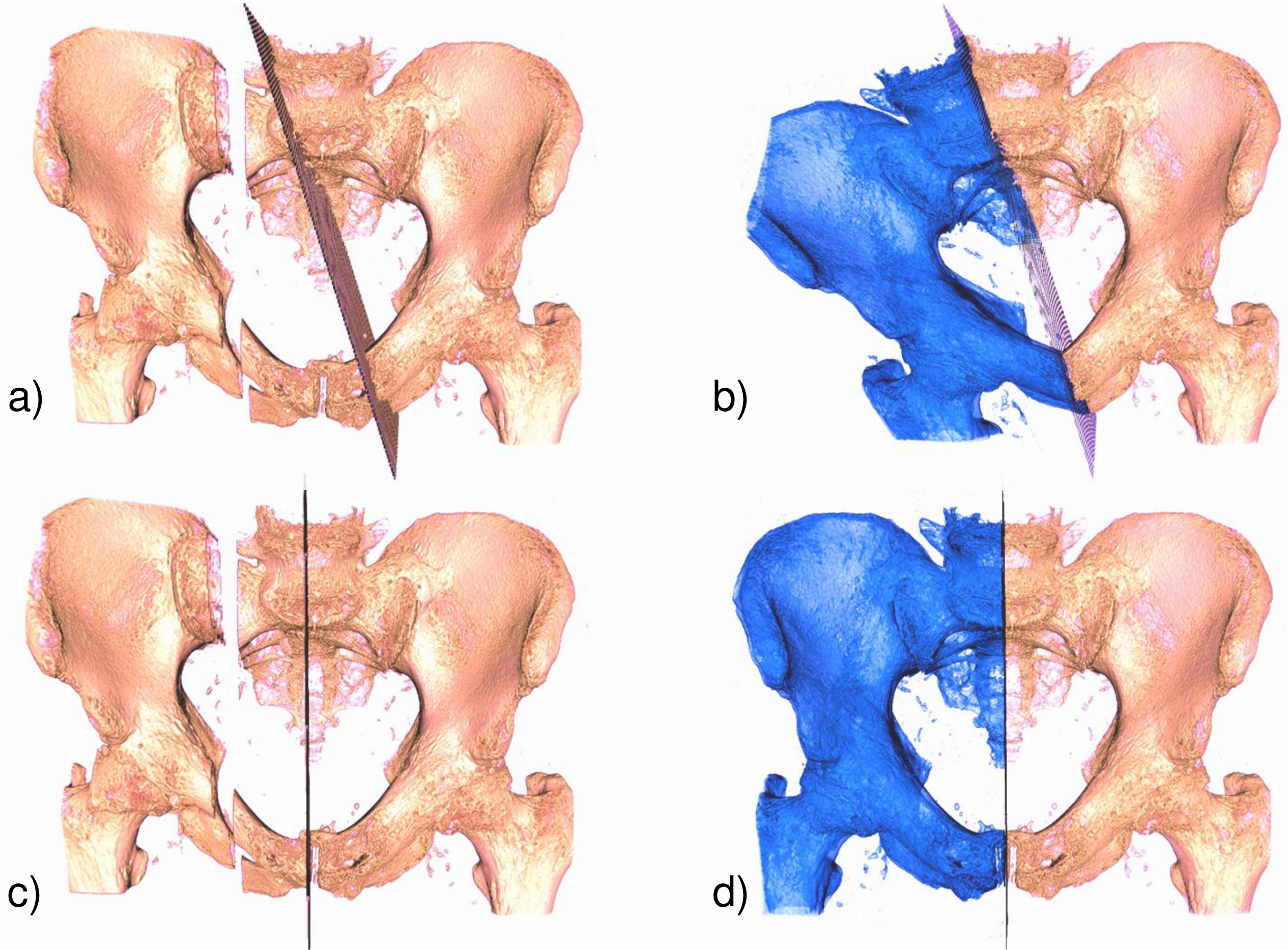}
	\caption{During an iterative strategy, the parameters associated to volumetric symmetry are estimated. In \textbf{(a-b)} the plane is visualized given the initial estimate, and in \textbf{(c-d)} it is visualized given the parameters at the convergence. The color blue represents the reconstructed bone model on the operative side of the patient.}
	\label{fig:iterations}
\end{figure}

Determining symmetry in 3D shapes has been a topic of interest in the field of computational geometry~\cite{mitra2013symmetry, xu2012multi}. Kazhdan et. al. proposed a Fourier-based descriptor to score reflective symmetries associated with planes passing through the center of the mass of the objects~\cite{kazhdan2002reflective,kazhdan2004reflective}. Podolak et. al. introduced an approach to recognize all symmetry planes, not limited to the ones passing through the center~\cite{podolak2006planar}. Symmetries were also detected in the form of intrinsic symmetries which includes all self-isometric deformations~\cite{ovsjanikov2008global, kim2010mobius}. Lastly, bilateral reflective symmetries were estimated within point clouds from the real environments using methods inspired from the iterative closes point strategy~\cite{combes2008automatic,ecins2017detecting,nagar2019detecting}.

\subsection{Clinical Motivation}~\label{subsec:clinicalMotiv}
Ensuring quality of fracture reduction in pelvis and acetabulum surgery is paramount. Studies have consistently shown that anatomic fracture reduction and stable fixation leads to improved outcomes in patients with these complex injuries~\cite{pastor2019quality,verbeek2017postoperative,pascarella2017surgical,shi2014radiographic,tornetta1996outcome,schenker2014pathogenesis}. For example, in a study of 31 patients who underwent open reduction internal fixation of an isolated pelvic fracture, Pastor et. al. demonstrated that clinical outcomes correlate with pelvis symmetry at a six-months time point~\cite{pastor2019quality}. Furthermore, studies demonstrate that anatomic articular surface reduction is critical in preventing post-traumatic arthritis in the acetabulum~\cite{giannoudis2010articular}.

However, ensuring adequate fracture reduction in pelvis and acetabulum surgery can be difficult. Most commonly 2D fluoroscopic C-arm imaging is used intra-operatively. Surgeons rely on known radiographic densities and parameters to guide fracture reduction. In acetabulum surgery, for example, surgeons acquire specific radiographic views to visualize the ilioischial and/or iliopectineal lines to assess reduction of the posterior and anterior pelvic columns respectively~\cite{blum2018vertical,mauffrey2018radiographic}. This method can become unreliable when the bone is comminuted and the anatomic area of interest is distorted. 

Additionally, surgeons may obtain radiographic views of the healthy contralateral side for comparison when attempting to reconstruct the comminuted operative side. Yet, this method has its own limitations. For instance, it is rare to obtain a radiographic view that possesses clear and symmetric views of both sides of the pelvis at one time. When the image intensifier is very close to the patient this may be achieved for the anterior/posterior (AP), inlet, and outlet views; however, many more radiographic views are typically used during a pelvis or acetabulum fracture case that does not possess this property. Additionally, even when both sides are visualized, the surgeon must imagine a mirrored version of the healthy side superimposed on the operative side, which increases the level of mental task load during the surgery. Studies have shown that these 2D fluoroscopic methods can be unreliable~\cite{lefaivre2014radiographic}, which has led surgeons to seek other methods for reduction assessment. 

Post-operatively, many surgeons acquire CT imaging to assess the quality of the reduction that was just performed, which has been shown to be superior to radiographic imaging~\cite{verbeek2017postoperative}. CT may also be used intra-operatively with cone-beam CT platforms. Yet these possess substantial drawbacks in terms of time, radiation, and change in surgical workflow. Additionally, these modalities can be cumbersome to use for the process of performing intra-operative fracture reduction as opposed to assessing the overall quality of a reduction after it is has been fully completed. 

Surgeons have also sought to take advantage of osseous symmetry for performance and assessment of fracture reduction. For example, Zhang et. al. 3D printed a mirrored model of the healthy side for comparison in lower limb long bone fractures~\cite{zhang2017can}. Symmetry has also been shown to be useful in the distal radius and facial fractures~\cite{gray2019image,bao2019quantitative}. However, neither viewing a 3D printed mirrored version of the healthy contralateral side, nor a mirrored radiographic image of the healthy side allow the surgeon to visualize the live comminuted operative side and the mirrored healthy side simultaneously and in the same position and orientation on the surgeon’s display. The ability to do so, could be substantially beneficial for intra-operative fracture reduction and assessment in pelvis and acetabulum fracture surgery. Similar concept can apply to other fields of surgery including brain and crainiofacial procedures~\cite{vannier1984three,raina2019exploiting, yu2009single,van1976pursuit,preuhs2019symmetry, liu2009symmetry}, or breast reconstruction procedures~\cite{edsander2001quality,nahabedian2005symmetrical,teo2018body}.

\subsection{Proposed Solution}~\label{subsec:proposedSolution}
In this work, we outline an end-to-end methodology to detect partial symmetry in human anatomy and exploit this knowledge intra-operatively as a reference to restore structural symmetry of fractured pelvic bone. Our fully automatic solution uses a structure-based cost, an intensity-based robust estimator, and a probabilistic-based loss to identify the plane of symmetry. After identifying the partial symmetry, healthy side of the patient anatomy is mirrored across the bilateral symmetry plane, which then allows simulating the ideal bone fragment configurations. This information is provided by overlaying the fluoroscopic image with a forward-projection of the mirrored anatomy obtained from the CT data. Our contributions enable the surgeon to use patient CT scans intra-operatively, without explicitly viewing the 3D data, but instead using 2D patient-specific image augmentation on commonly used X-ray images. In contrast to surgical navigation systems that provide update in relation to pre-operative data, our solution provides interventional feedback with respect to the desired outcome.

Since we introduced our preliminary work on using symmetry for pelvic trauma cases (\cite{fotouhi2018exploiting}), there have been multiple other works that investigated the usability of this approach and quantified the extent to which human pelvis is symmetric, and verified this hypothesis~\cite{de2019regional, ead2020investigation,ead2020virtual}. The present manuscript substantially extends our preliminary work by substituting the manual initialization step with an automatic initialization strategy via intrinsic symmetry, extracting the subset of extrinsic symmetry-invariant features from intrinsic symmetry-invariant features with a multi-stage RANSAC, and parameterizing extrinsic symmetry based on 3D feature points. We also present the discussion regarding the robustness of the Tukey-based similarity cost for detection of contra-lateral symmetry, and extended discussion on the regularization measure using mutual information. Furthermore, we present new experimental results by including the evaluation against different regularization factors, evaluation regarding the robustness to noise and the outlier extent (imperfect symmetry), quantification of the capture range in the optimization landscape, comparison between the outcome of our proposed automatic initialization strategy vs. the problem’s capture range, error estimation in detecting symmetry in pelvis data, and outcome on real patient data with severe pelvic injuries.

\section{Materials and Method}

\subsection{Theory}
Bilateral reflective symmetry is defined as the group of involutive isometric maps $\mat{M}_g \in \bar{E}(3)$, where $\bar{E}(3)$ consists of self-isometries such that $\bar{E}(3) = \{ h \in E(3) , \mat{o} \subseteq \mathbb{P}^3 \enspace \vert \enspace h (\mat{o}) = \mat{o} \}$, where $\mat{o}$ is defined in the 3D projective space $\mathbb{P}^3$. The group $E(3)$ denotes all isometries of $\mathbb{R}^3$. Transformation $\mat{M}_g$ mirrors the object $\mat{o}$ across a symmetry plane such that ${\mat{o}}^{-} = \mat{M}_g ({\mat{o}}^{+})$, where $\mat{o}^{-} , \mat{o}^{+} \subseteq \mathbb{P}^3$ are sub-volumes of object $\mat{o}$ on the opposite sides of the symmetry plane.  

Assuming the plane of symmetry is the Y-Z plane, the extrinsic symmetry is expressed via  $\mat{M}_g := g \, m_x \, g^{-1}$, where $g$ is a member of the Special Euclidean group  $SE(3)$, and $m_x$ reflects the space about the X-axis:
\begin{equation}
    m_x = 
    \begin{bmatrix} 
    -1 & 0 & 0 & 0 \\
    0 & 1 & 0 & 0 \\
    0 & 0 & 1 & 0 \\
    0 & 0 & 0 & 1 \\
    \end{bmatrix}.
\end{equation}

Transformation $\mat{M}_g$ maps the points $\mat{p},\mat{q} \in \mathbb{P}^3$ to $\bar{\mat{p}} = g \, m_x \, g^{-1} \, \mat{p}$ and $\bar{\mat{q}} = g \, m_x \, g^{-1} \, \mat{q}$, respectively. The distance between $\bar{\mat{p}}$ and $\bar{\mat{q}}$ is then computed as: 
\begin{equation}
    \left\lVert \bar{\mat{p}} - \bar{\mat{q}} \right\rVert_2 = 
    \left\lVert g \, m_x \, g^{-1} \, (\mat{p - q}) \right\rVert_2.
\end{equation}
Since, $ \text{det}(g \, m_x \, g^{-1})  =  \text{det}(g) (-1) (\text{det}(g))^{-1} = -1$, then \textit{i)} $\left\lVert \bar{\mat{p}} - \bar{\mat{q}} \right\rVert_2 = \left\lVert \mat{p} - \mat{q} \right\rVert_2$, hence $\mat{M}_g$ is an isometry, and \textit{ii)} due to the negative determinant, $\mat{M}_g$ is orientation reversing, therefore is an anti-conformal map.

\subsection{Problem Formulation}\label{seubsec:probFormul}
We propose to estimate extrinsic symmetry parametrization by minimizing the following cost:
\begin{equation}\label{eq:cost}
\argmin_{g} D(\mat{M}_g | g_0) := d_{I}(\mat{o}, \mat{M}_g(\mat{o})) + \lambda \; d_{D}(\mat{o}, \mat{M}_g(\mat{o})).
\end{equation}

In Sec.~\ref{subsec:Mobius} the prior parametrization $g_0$ is automatically estimated based on surface correspondences detected from the anatomy. In Sec.~\ref{subsec:TuK} we present a robust estimator term that will be used to minimize an intensity-based distance $d_I(.)$, followed by a distribution-based regularization term $d_D(.)$ which will be discussed in Sec.~\ref{subsec:MI}. Fig.~\ref{fig:iterations} demonstrates the iterative step that yields the optimal plane of bilateral symmetry by minimizing the total loss $D(.)$. Finally, surgical image augmentation using interventional image registration is discussed in Sec.~\ref{subsec:AR}.


\subsection{Automatic Initialization of Extrinsic Symmetry via Global Intrinsic Symmetry}\label{subsec:Mobius}
\begin{figure}
	\centering
		\includegraphics[width=\columnwidth]{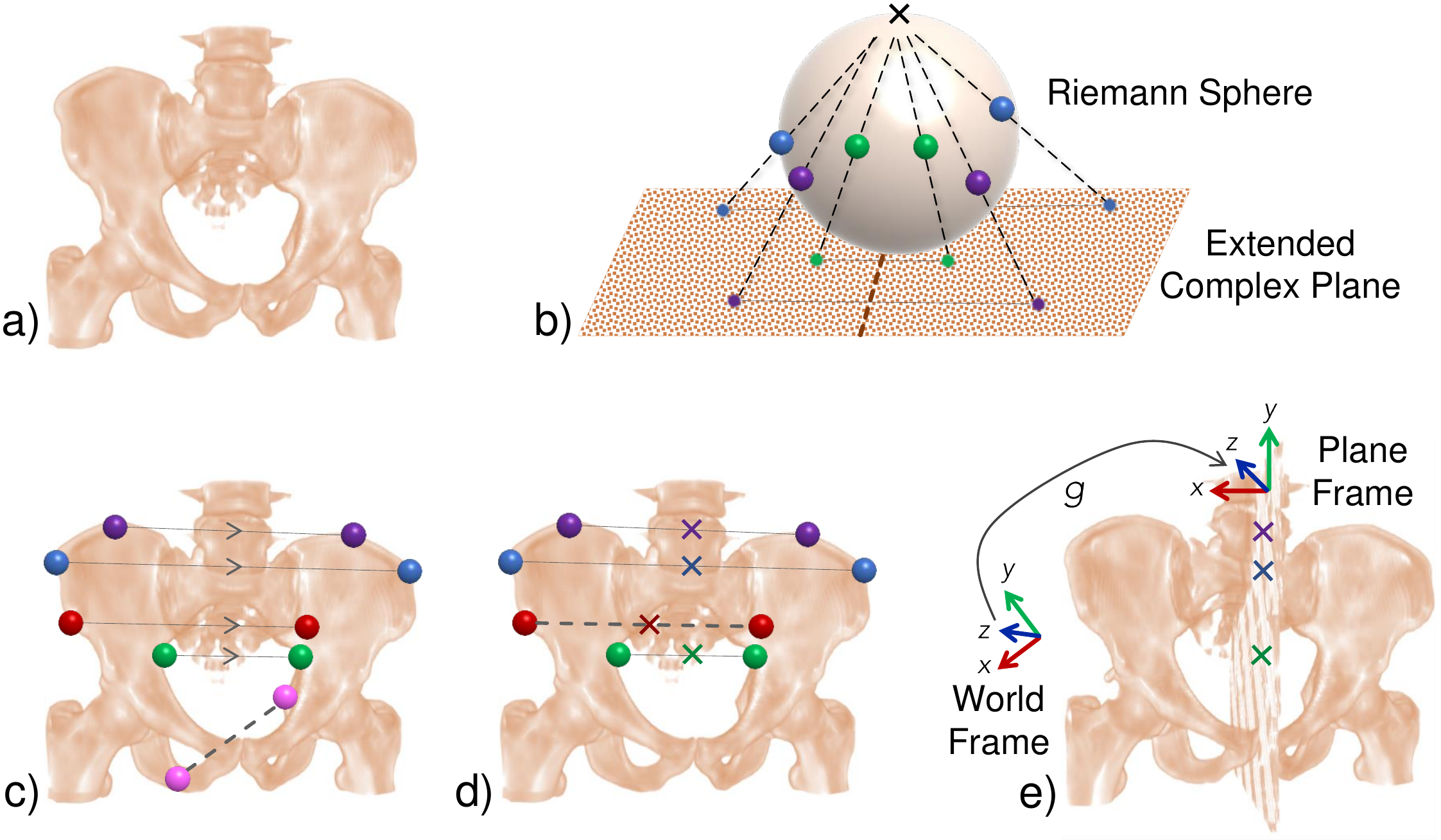}
	\caption{First step in computing the initial parametrization of extrinsic symmetry on an object that exhibits imperfect symmetry involves transforming the surface of the object to the Reimann sphere (uniformization) and consequently to the extended complex plane (stereographic projection). These transformations are shown in parts \textbf{(a-b)}. Anti-M\"obius group is then used to model the intrinsic symmetry on the complex plane and assign self-symmetry correspondences to vertices on the surface of the pelvis. During a two-step RANSAC scheme, the correspondences are pruned and narrowed down to only inliers which satisfy the extrinsic symmetry properties \textbf{(c-d)}. As the result of the first and second RANSAC, correspondeces shown in dashed lines are removed due to dissimilar directions and midpoints, respectively. Extrinsic symmetry is subsequently computed using least squares minimization \textbf{(e)}.}
	\label{fig:MobiusPipeline}
\end{figure}

To compute an initial estimate $g_0$ for the plane of partial symmetry, we first estimate a set of point correspondences on the contralateral sides of the anatomy which satisfy the properties of intrinsic symmetry. Intrinsic symmetry $\mat{M}_i(.)$ is associated with all volumetric deformations that preserve pairwise geodesic distances on a symmetric surface.  All correspondences are detected automatically and globally~\cite{kim2010mobius}, and are used to compute an initial estimate of the partial symmetry plane. An overview of this step is demonstrated in Fig~\ref{fig:MobiusPipeline}.

\subsubsection{Symmetry Invariant Candidates} The candidate symmetry invariant correspondences $p$ are selected such that they share a common intrinsic symmetry, \textit{i.e.} $\mat{M}_i(p) = p$. Selecting symmetry invariant point candidates are particularly crucial in the pelvic trauma application due to the imperfect and incomplete symmetry which are caused by fractures and dislocations. 

The critical points $p$ of a symmetry invariant function $\Phi = \{\Phi(.)\!: \mat{o} \rightarrow \mathbb{R}$, $\Phi(\mat{M}_i(p)) = \Phi(p)\}$, are invariant to symmetry on the surface of the object $\mat{o}$. To verify this, from chain rule, we drive:
\begin{equation}
    \nabla \Phi(\mat{M}_i(p)) \; \mat{M}_i'(p) = \nabla \Phi(p).
\end{equation}
For the critical point $p_c$, it can be shown that $\nabla \Phi(\mat{M}_i(p_c)) = 0$ \textit{if-and-only-if} $\nabla \Phi(p_c) = 0$ , implying that $\mat{M}_i(p_c) = p_c$, hence $p_c$ satisfying symmetry invariance condition. 

The Average Geodesic Distance function is used as the symmetry invariant function to generate candidate points and is defined as below~\cite{kim2010mobius}:
\begin{equation}
    \Phi(p) = \int_{q \in \mat{o}} d(p,q) \, dq.
\end{equation}

\subsubsection{Optimal Intrinsic Symmetry via M\"obius Transform}
Every genus zero surface can be mapped to a unit sphere $\mathbb{S}$, also known as the Riemann sphere~\cite{pinkall1993computing}. The group of M\"obius transformations models all the angle preserving isometries between the Riemann sphere to itself. Via stereographic projection, Riemann sphere can then be mapped to the extended complex plane. Therefore, M\"obius transformation is represented as the mapping between extended complex planes. M\"obius transformation is formulated on the extended plane via the fractional linear function, also known as homographies:
\begin{equation}
    h(z) = \frac{az + b}{cz + d}\,, \qquad a,b,c,d \in \mathbb{C}.
\end{equation}
M\"obius group that models all isometries is characterized using only $6$ real parameters, hence $3$ point correspondences on the complex plane are sufficient to uniquely compute a M\"obius transformation in a closed form. This property allows the intrinsic symmetry on a mesh to appear as extrinsic symmetry on the extended complex plane. Using only $6$ parameters has the advantage of computing the parametrization for intrinsic symmetry both globally and efficiently. 

The search for the parametrization of symmetry takes place on the anti-M\"obius group, which includes the M\"obius group augmented with reflections~\cite{kim2010mobius}. The anti-M\"obius transforms cover the set of orientation reversing isometries which are instances of the anti-conformal maps.

To use the M\"obius parameterization, we first construct a binarized volume by segmenting bone from tissue and air using automatic histogram thresholding using the appropriate HU ranges for each tissue. 
Since the focus of this step is on autonomy for initialization, we avoided interactive or complex segmentations that may require adjustments or parameter tuning. The minor sensitivity to varying bone appearances during segmentation can be compensated later by multiple outlier rejection steps built into our pipeline, which will be introduced later in the manuscript.
Once the segmentation is achieved, a genus zero surface is constructed from the object using Reeb graphs~\cite{dey2013reeb}. The pseudo-code for this step is presented in Alg.~\ref{alg:reeb}.

\begin{algorithm}
\caption{Construction of the genus zero mesh}\label{alg:reeb}
\begin{algorithmic}[1]
\State initialize empty graph $G=(V=\{\}, E=\{\})$
\State initialize empty vertex set $V_{\text{previous}}=\{\}$
\item[]
    \For {each transverse slice $s=0,1,2,\dots$}
        \State initialize empty vertex set $V_{\text{current}}=\{\}$
        \For{each connected component $c=0,1,2,\dots$}
            \State add vertex with tag $(s, c)$ into $V$ and $V_{\text{current}}$
            \For{each element $(s_p, c_p)$ in $V_{\text{previous}}$}
                \If {$(s, c)$ is connected to $(s_p, c_p)$}
                    \State add edge $(s,c)-(s_p,c_p)$ to $E$
                \EndIf
            \EndFor
            \State $V_{\text{previous}} = V_{\text{current}}$
        \EndFor
    \EndFor
\item[]
\State close all cycles in the Reeb graph $G$ by adding the convex hull of the connected components
\end{algorithmic}
\end{algorithm}

On the genus zero surface, a set of feature points (triplets and quadruplets of point correspondences) are selected iteratively, and each time the parameters of a unique M\"obius transformation is estimated. Followed by each step, all surface elements on the complex plane are warped given the current M\"obius estimate, and pair-wise geodesics are measured. The mapping that yields the most number of mutually closest points (most inliers) is selected to parametrize the intrinsic symmetry on the surface~\cite{lipman2009mobius}. 

\subsubsection{Refinement of Point Correspondences}
To incorporate the most reliable candidates for estimating the parameters of extrinsic symmetry from the set of point correspondences, we employ a two-stage RANdom SAmple Consensus (RANSAC) strategy~\cite{fischler1981random}. After the completion of this step, the outlier correspondences that are not with agreement with the global symmetry properties are eliminated from the list. 

The vector connecting a point $p$ on the surface to its reflection $p' = g m_x g^{-1} p$ is given by $(g m_x g^{-1} - \mathbb{1})p$. For any arbitrary point $p$, it can be shown that this vector is parallel to $\mat{g}_1$, where $g = \begin{bmatrix} \mat{g}_1 & \mat{g}_2 & \mat{g}_3 & \mat{g}_4 \end{bmatrix}$. This equality can be intuitively explained given the convention that the plane normal is parallel to the X-axis of the local frame. Therefore, bilateral reflections across the symmetry plane occur in the direction of the local X-axis. The local X-axis of the plane is parallel to $\mat{g}_1$ which expresses the image of the world X-axis: $g_1 = g\begin{bmatrix}1 & 0 & 0 & 0 \end{bmatrix}^{\top}$. Hence, the vectors of all correspondences that satisfy extrinsic symmetry are jointly parallel to $g_1$. We leverage these relations in a RANSAC setting and eliminate all correspondences with non-agreeing connecting vectors.

During a second RANSAC, we seek consensus among the midpoints of the correspondences. The midpoint $\bar{p}$ lies on the symmetry plane and is invariant to reflection:
\begin{equation}
\begin{aligned}
     \mat{M}_g \, \bar{p} & = g m_x g^{-1} (\frac{p + p'}{2}) = g m_x g^{-1}(\frac{p +  g m_x g^{-1}p}{2})\\
     & = g m_x g^{-1}(\frac{\mathbb{I} + g m_x g^{-1}}{2}p)= \frac{g m_x g^{-1} + \mathbb{I}}{2}p = \bar{p}.
\end{aligned}
\end{equation}
We employ the second RANSAC to remove outliers from the set of correspondences where there are no strong agreements on the midpoints. At the end of this stage, we identify the key points that only satisfy extrinsic symmetry, selected from a subset of the correspondences that satisfied intrinsic symmetry on the surface.

\subsubsection{Extrinsic Symmetry from Point Correspondences}
The parametrization $g_0$ of the partial symmetry plane in 3D is defined via a point $\mat{n}_0$ and a plane normal $\vec{\mat{n}}$. To obtain these parameters which describe extrinsic symmetry, we compute the mean-normalized matrix $\bar{P}$ of all $N$ midpoints. Singular value decomposition of this matrix yields:
\begin{equation}
    \bar{{P}}_{n \times 3} = U_{n \times n} \Sigma_{n \times 3} V^{T}_{3 \times 3}.
\end{equation}
The last column vector $\mat{v}_3$ of $V = \begin{bmatrix} \mat{v}_1 \; \mat{v}_2 \; \mat{v}_3 \end{bmatrix}$ defines the least principle component of the data, \textit{i.e.} axis with minimum variance. This vector defines the normal to a plane that has the closest distance to all midpoints. Lastly, the point $\mat{n}_0$ is computed as the mean of all midpoints: $E[\bar{\mat{p}}]$. 


\subsection{Robust Estimators for Detecting Imperfect Symmetry}\label{subsec:TuK}

After an initial parametrization $g_0$ of the plane is obtained, the intensities are compared iteratively between the voxel elements across the plane of partial symmetry, until the parameters $\mat{M}_g$ that minimize the intensity loss $d_{I}(\mat{o}, \mat{M}_g(\mat{o}))$ are estimated.  

A major challenge associated with using an intensity-based approach is the presence of severe outlier regions which may result in incorrect symmetry parametrization given an intensity-based criteria. The symmetry outliers are the results of \textit{i)} unilateral dislocations and fractures, and \textit{ii)} imperfect symmetry in the original anatomy. Normalized Cross-Correlation (NCC) is a commonly used intensity-based measure which is greatly sensitive to noise and outliers~\cite{fotouhi2018exploiting}. Therefore, it is crucial to employ a symmetry detection mechanism that is robust to the outlier regions. We suggest using Tukey biweight robust estimator which automatically downweights or supresses the regions that exhibit consistently high errors, and prohibits those elements from contributing to the total loss~\cite{huber2011robust}. Inspired by Tukey robust regression we suggest the following loss:
\begin{equation}
    d_{I}(\mat{o}, \mat{M}_g(\mat{o})) = \sum_{i=1}^{ |\Omega_s|} \frac{| \rho (e_i(\mat{M}_g))|}{|\Omega_s|},
\end{equation}
where $\Omega_s$ is the spatial domain of the volumetric data. The element-wise error $\rho (e_i(\mat{M}_g))$ is computed as follows:
\begin{equation}
    \rho (e_i(\mat{M}_g)) = 
    \begin{cases}
	    e_i(\mat{M}_g)\left[ 1 - \left( \frac{e_i(\mat{M}_g)}{c} \right)^2 \right]^2 &;  |e_i(\mat{M}_g)| \leqslant c,\\
        0 &;  \text{otherwise}.
    \end{cases}
\end{equation}
Parameter $c$ sets a threshold that is used for classifying the voxel elements as outliers. The element-wise weighted residuals are computed as:
\begin{equation}\label{eq:e(M)}
        e_i(\mat{M}_g) = \frac{ r_i(\mat{M}_g)}{S} \, , \text{such that}  \ \,  r_i(\mat{M}_g) = \mathcal{I}(o_{i}) - \mathcal{I}(\mat{M}_g (o_{i})).
\end{equation}
In Eq.~\ref{eq:e(M)}, the term $\mathcal{I}(.)$ denotes voxel intensity. 

It is suggested in the literature that $c = 4.685$ provides around $95\%$ asymptotic efficiency of linear regression for normal distributions~\cite{huber2011robust}. This value is computed assuming the residuals $e_i$ are drawn from a unit variance distribution. To relax this constraint, the factor S is used as a scaling parameter and is computed based on median absolute deviation:
\begin{equation}
    S = \frac{MD}{0.6745} \, , \quad MD = \underset{j \in \Omega_s}{\median(r_j)}.
\end{equation}

\begin{figure}[ht!]
	\centering
		\includegraphics[width=\columnwidth]{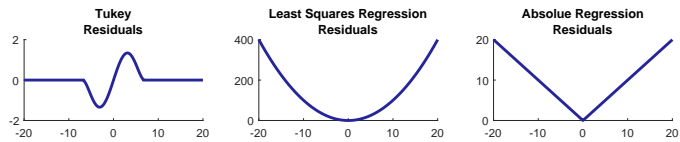}
	\caption{Comparison of the Tukey bi-weight with L2 and L1 norms. The horizontal axis represents the residual error, and the vertical axis shows the corresponding loss.}
	\label{fig:tukeyweights}
\end{figure}

As demonstrated in Fig.~\ref{fig:tukeyweights}, in contrast to L2 and L1 norms, Tukey biweight disparity term completely suppresses the residuals beyond a threshold, regarding the elements yielding excessive errors as outliers. This behaviour is desired for traumatic cases, since the outlier regions, \textit{i.e.} dislocated bone, should be completely excluded during symmetry identification, allowing the plane parameter estimation to solely rely on the partial symmetry present in the anatomy. In our previous work, we presented a comparison between Tukey biweight loss and normalized cross-correlation (NCC) that is commonly used in medical image registration~\cite{fotouhi2018exploiting}. Results favored Tukey loss over NCC for the cases with outliers and noise.

\subsection{Regularization based on Bone Distribution}\label{subsec:MI}
\begin{figure}[ht!]
	\centering
		\includegraphics[width=\columnwidth]{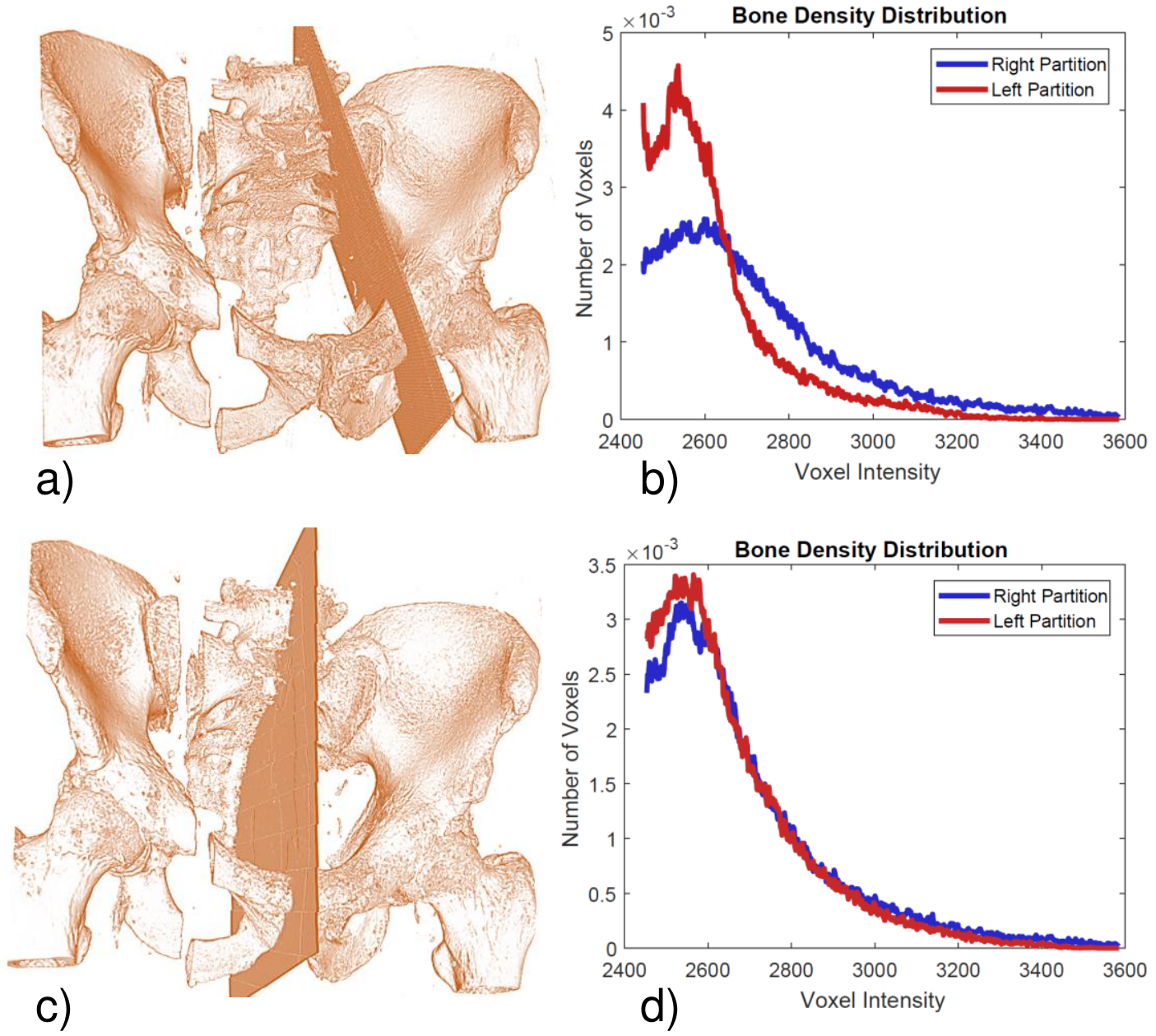}
	\caption{Distribution of the bone HU values across the symmetry plane before and after the estimation of symmetry plane. Comparing the histograms in (b) and (d) suggests high similarity when the plane dissect the volume bilaterally in the center.}
	\label{fig:histogram}
\end{figure}

To further support the identification of partial symmetry, we exploit a biological fact that, even if injured, dislocated bone fragments remain within the body. On this basis, a regularization term is introduced to maximizes the similarity between the distribution of bone densities across the symmetry plane. This concept is visualized in Fig.~\ref{fig:histogram}.

To materialize this notion, probability distribution functions are computed in the form of image histograms from voxel intensities. A regularizer based on Normalized Mutual Information (NMI) is used to acquire a similarity score between the density distributions on the contralateral sides:
\begin{equation}\label{eq:NMI}
d_{D}(\mat{o}, \mat{M}_g (\mat{o})) = - \frac{H\Big(\mathcal{I}(\mat{o})\Big) + H\Big(\mathcal{I}(\mat{M}_g (\mat{o}))\Big)}{H\Big(\mathcal{I}(\mat{o}), \mathcal{I}(\mat{M}_g (\mat{o}))\Big)}.
\end{equation}
In the formulation presented in Eq.~\ref{eq:NMI}, $H(.)$ is the entropy of voxels' intensities. 

\begin{figure}[ht!]
	\centering
		\includegraphics[width=\columnwidth]{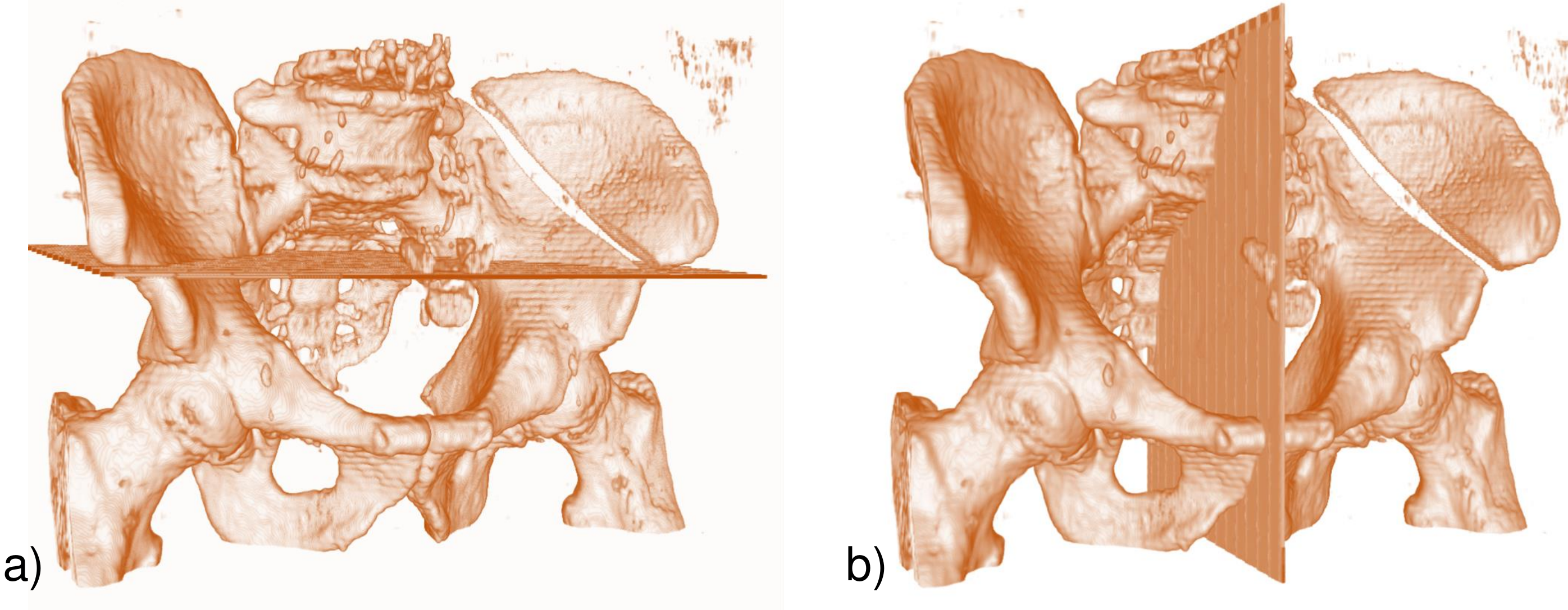}
	\caption{Two different plane estimates with near identical distribution score, and substantially different Tukey-based score.}
	\label{fig:tuK_mi_good_bad}
\end{figure}

The regularizer term $d_D (.)$ is globally non-convex and yields local minimas for various plane parametrizations. An example case is shown in Fig.~\ref{fig:tuK_mi_good_bad}, where the NMI-based score is nearly equal for two cases, while the robust estimator cost measured based on Tukey biweight disparity is substantially ($99.1\%$) lower at the configuration in Fig.~\ref{fig:tuK_mi_good_bad}-b. Therefore, the density-based cost cannot replace the Tukey-based term, and is merely used as a regularizer to ensure similar bone distributions contralaterally. 

\subsection{Interventional Image Registration and Augmentation}\label{subsec:AR}
The plane of partial symmetry with the parametrization $\mat{M}_g$ dissects the volume bilaterally, yielding sub-volumes $\mat{o}_i$ and $\mat{o}_h$, denoting the injured and healthy regions, respectively ($\mat{o} = [\mat{o}_i, \mat{o}_h]$). Using this parametrization, the non-fractured portion of the data can be mirrored across the plane as $\bar{\mat{o}}_h = \mat{M}_g(\mat{o}_h)$, resulting in a non-fractured model of the patient: $\bar{\mat{o}} = [\bar{\mat{o}}_h, \mat{o}_h]$. This patient-specific reconstructed model is then used as a template of patient anatomy, representing the anatomical structures \textit{"as if they were repaired"}. It is important to stress that, although the human pelvic skeleton is not entirely symmetric, it is common for an orthopedic trauma surgeon to consider it symmetric, and use the contralateral side as reference.

\begin{figure}
	\centering
		\includegraphics[width=\columnwidth]{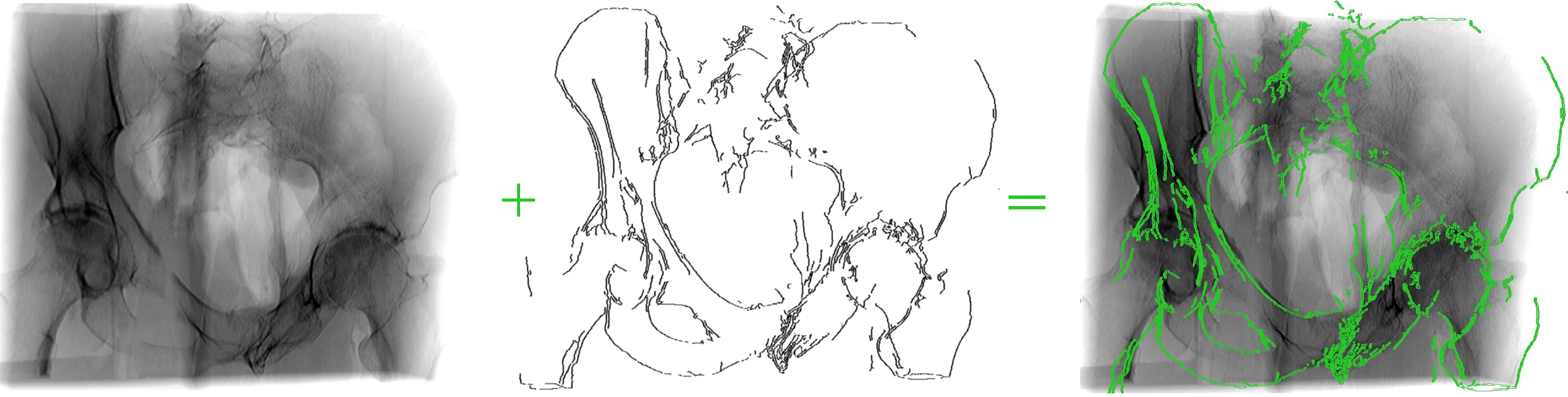}
	\caption{Interventional X-ray images are augmented with the contours of the bone extracted from the mirrored CT volume. The green contours serve as road-map, demonstrating desired configurations for bone fragments in the perspective of each X-ray image such that bilateral symmetry is restored.}
	\label{fig:overlay}
\end{figure}

To exploit the patient-specific template $\bar{\mat{o}}$ intra-operatively, each interventional C-arm fluoroscopy image is augmented with the contours of the reconstructed bone in the mirrored CT volume as demonstrated in Fig.~\ref{fig:overlay}. Enabling such augmentation requires two steps. In the first step, the transformation that describes the projective relation between the pre-operative CT image and the intra-operative X-ray image is computed via 2D/3D image registration by maximizing NCC score defined below: 

\begin{equation}
    \argmax_{R, \mat{t}, k} \, NCC\left( R, \mat{t}, k | \mathcal{I}_{X}, \mathcal{I}_{D} \right) = \sum^{| \Omega_{X , D} |} \frac{\mathcal{I}_{X} \cdot \mathcal{I}_{D}( R, \mat{t}, k \big| \mat{o})  }{ \sigma_{X} \sigma_{D} }.
\end{equation}
This formulation optimizes over the parameters of the rotation $R$, translation $\mat{t}$, and intrinsic geometry $k$. The parameters $\mathcal{I}_{X}$ and $\mathcal{I}_{D}$ are the mean-normalized X-ray and Digitally Reconstructed Radiographs (DRRs) generated from the fractured patient data given the parameters $(R, \mat{t}, k)$. Finally, $\Omega_{X , D}$ is the common spatial domain of the two images, and $\sigma_{X}$ and $\sigma_{D}$ are the standard deviations of the X-ray and DRR images within $\Omega_{X, D}$. In the second step, augmented image $\mathcal{I}_A$ is constructed by overlaying the X-ray image and the 2D DRR that is computed from the mirrored volume, \textit{i.e.} $\mathcal{I}_A := \mathcal{I}_X \, \cup \, \mathcal{I}_{D}( R, \mat{t}, k , \bar{\mat{o}})$.


\section{Experimental Results}\label{sec:results}
\begin{figure}
	\centering
		\includegraphics[width=\columnwidth]{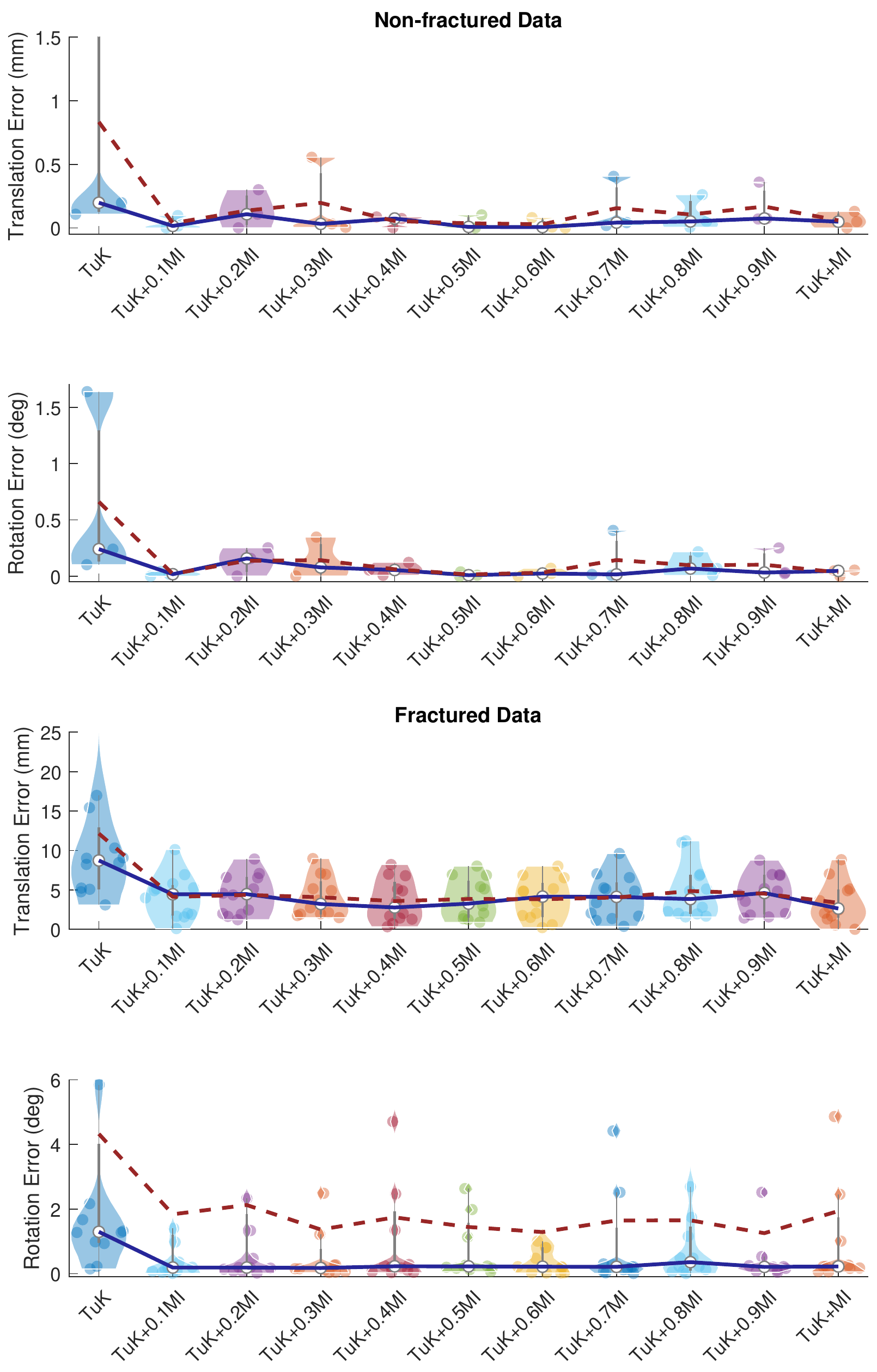}
	\caption{Translation and rotation errors are shown for different regularization factor $\lambda \in [0, 1]$. The solid blue lines represent the median, and the dashed red lines depict the mean values.}
	\label{fig:TuKMI}
\end{figure}
In this section we evaluate the proposed methodology for detecting and exploiting symmetry under a variety of different conditions. We present both quantitative and qualitative outcome on data with synthetic fractures, as well as patient data with complex unilateral fractures. To generate ground-truth for the evaluations in  sections~\ref{subsec:1}-\ref{subsec:4}, we synthetically construct symmetric pelvis data using patient cases from the NIH Cancer Imaging Archive~\cite{clark2013cancer}. In Sec.~\ref{subsec:5}, the distance between anatomical landmarks on pelvic cases with simulated unilateral fractures were compared before and after applying the symmetry transform. Finally in Sec.~\ref{subsec:6}, the symmetry detection and surgical image augmentation is demonstrated on three patient data with unilateral pelvic fractures. 

Bound constrained by quadratic approximation method~\cite{powell2009bobyqa} was used for the optimization of the non-linear cost in Eq.~\ref{eq:cost}. For all experiments presented in this section, the maximum number of iterations was set to $100$. The rotation errors are measured as the angle between the normal $\hat{\mat{r}}$ of the estimated symmetry plane and the ground-truth normal $\mat{r}_1$. Since the plane normal is in the direction of the X-axis of the local frame, hence the normal vectors are defined via the first column vector of the rotation matrix associated to the plane pose. The rotation error is formulated as:
\begin{equation}
    \theta_e = \cos^{-1} (\hat{\mat{r}}_1 ^{\top} \mat{r}_1).
\end{equation}
Since Euclidean distance is not explicitly defined for non-parallel planes, we define a translation measure as the projection of the translation difference $(\mat{t} - \hat{\mat{t}})$ onto the plane normal:
\begin{equation}
    \mat{t}_e = \left| (\mat{t} - \hat{\mat{t}})^{\top} \mat{r}_1 \right|. 
\end{equation}

\begin{figure}
	\centering
		\includegraphics[width=\columnwidth]{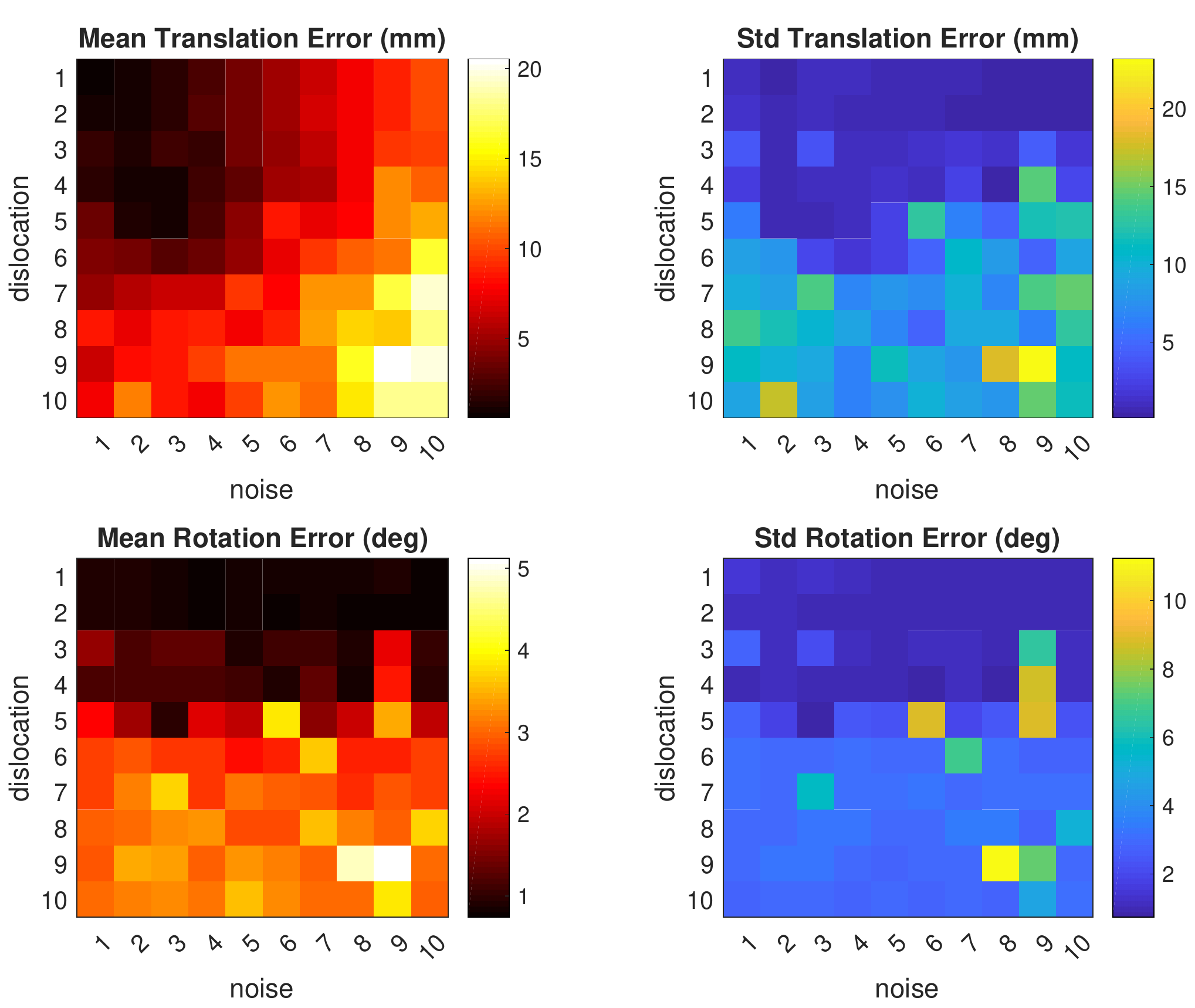}
	\caption{Error in detection of partial symmetry plane is evaluated against different amounts of noise and bone dislocation. It should be noted that the color scale for the heatmaps are different to better illustrate the differences within each figure.}
	\label{fig:noisedislocation}
\end{figure}

\subsection{On the Effect of Regularization}\label{subsec:1}
We evaluated the performance of the suggested symmetry detection cost in Eq.~\ref{eq:cost} at different regularization levels for ten cases, where $\lambda$ was varied between $0.0$ and $1.0$. In Fig.~\ref{fig:TuKMI} results are presented for data without fractures as well as data with different fracture patterns. Each case is evaluated $4$ times, each with a random initialization in a neighborhood around the ground-truth within the ranges of $\pm 15$\,mm and $\pm 15^{\circ}$ for translation and rotation along each axis, respectively. As shown in the results, all regularizers improve the convergence compared to the case where no regularization is used, however no significant differences is observed between different regularization parameters.

\subsection{Sensitivity to Imperfect Symmetry}\label{subsec:2}
The performance of the symmetry cost $D(.)$ presented in Eq.~\ref{eq:cost} was tested against different levels of noise and bone dislocations. In all experiments, the regularization factor $\lambda$ was set to $0.5$ for consistency, allowing $d_I(.)$ to be the dominant term driving the total cost, and $d_d(.)$ serving as a fidelity term. The amount of dislocation (outlier) was varied between $0\%$ to $30\%$ of the entire volume, and the Gaussian noise between $0\%$ and $40\%$ of the highest intensity in the volume. For each given outlier and noise level, the symmetry detection was repeated 20 time, each time randomly sampling an initialization parameter within the maximum range of $\pm 15$\,mm and $\pm 15^{\circ}$ around the ground-truth. Results are presented in Fig.~\ref{fig:noisedislocation}.

\subsection{Capture Range}\label{subsec:3}

\begin{figure}
	\centering
		\includegraphics[width=0.9\columnwidth]{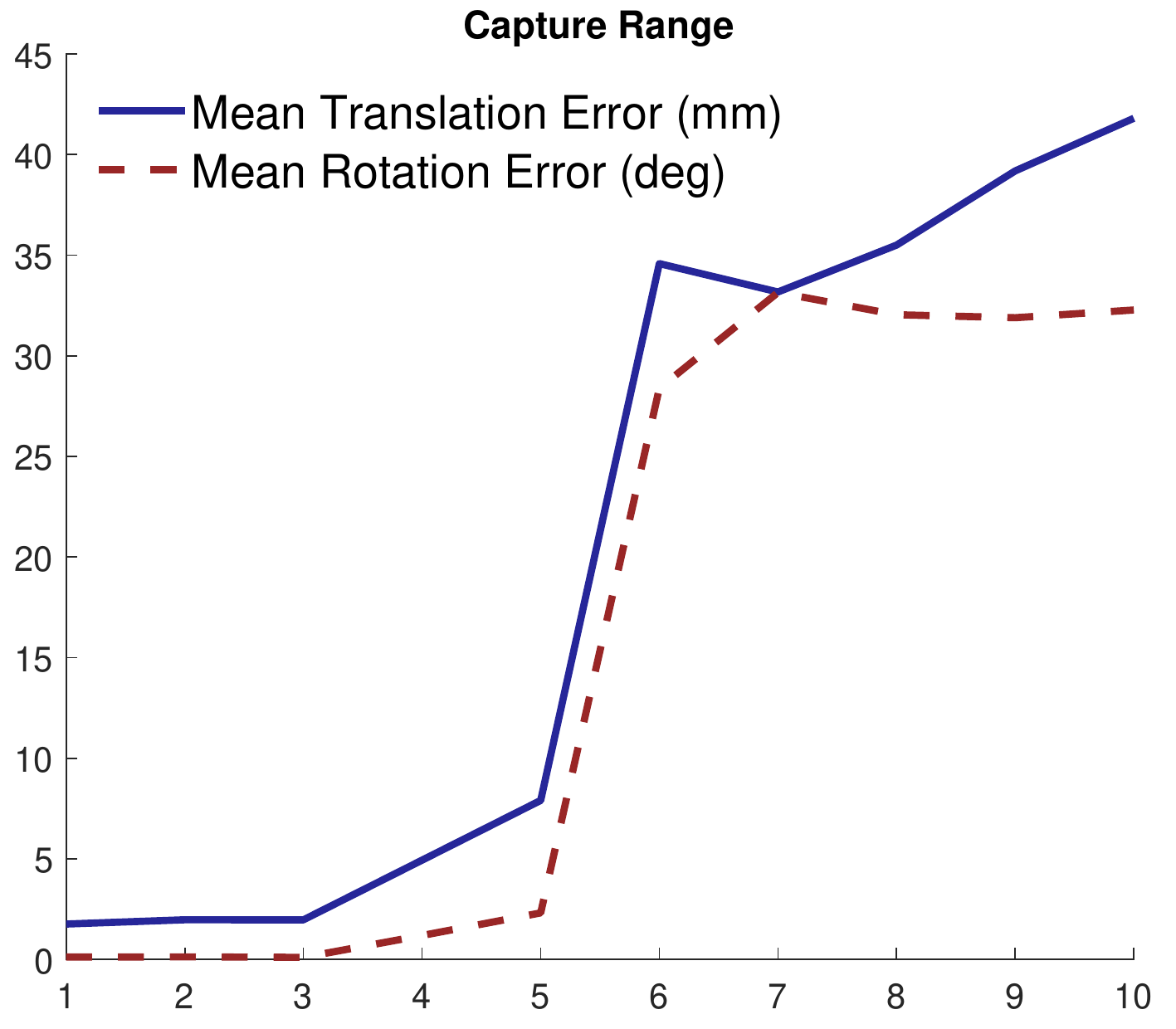}
	\caption{Dependence of the regularized Tukey cost on initialization. The horizontal axis corresponds to the extent of misalignment at the intialization step, and the vertical axis represents the translation and rotational errors after convergence.}
	\label{fig:captureRange}
\end{figure}

We characterize the dependence of the regularized Tukey cost on the initialization parameters. In Fig.~\ref{fig:captureRange}, the mean rotation and translation errors are presented for varying initialization parameters. The elements on the horizontal axis represent the level of misalignment at the initial configuration, where at each step the ranges for the initialization misalignment are increased by $5$\,mm and $5^{\circ}$ along each axis of translation and rotation, respectively. For instance, the first and second elements on the horizontal axis in Fig.~\ref{fig:captureRange} which have yielded lower errors show initialization samples for each axis between the ranges of $\left[ (0\,\text{mm}, 5\,\text{mm})-(0^{\circ}, 5^{\circ})\right]$ and $\left[(5\,\text{mm}, 10\,\text{mm})-(5^{\circ} , 10^{\circ})\right]$, respectively. Given each range, the sampling is repeated $10$ times. 

\subsection{Accuracy of Global Initialization}\label{subsec:4}
\begin{figure}
	\centering
		\includegraphics[width=\columnwidth]{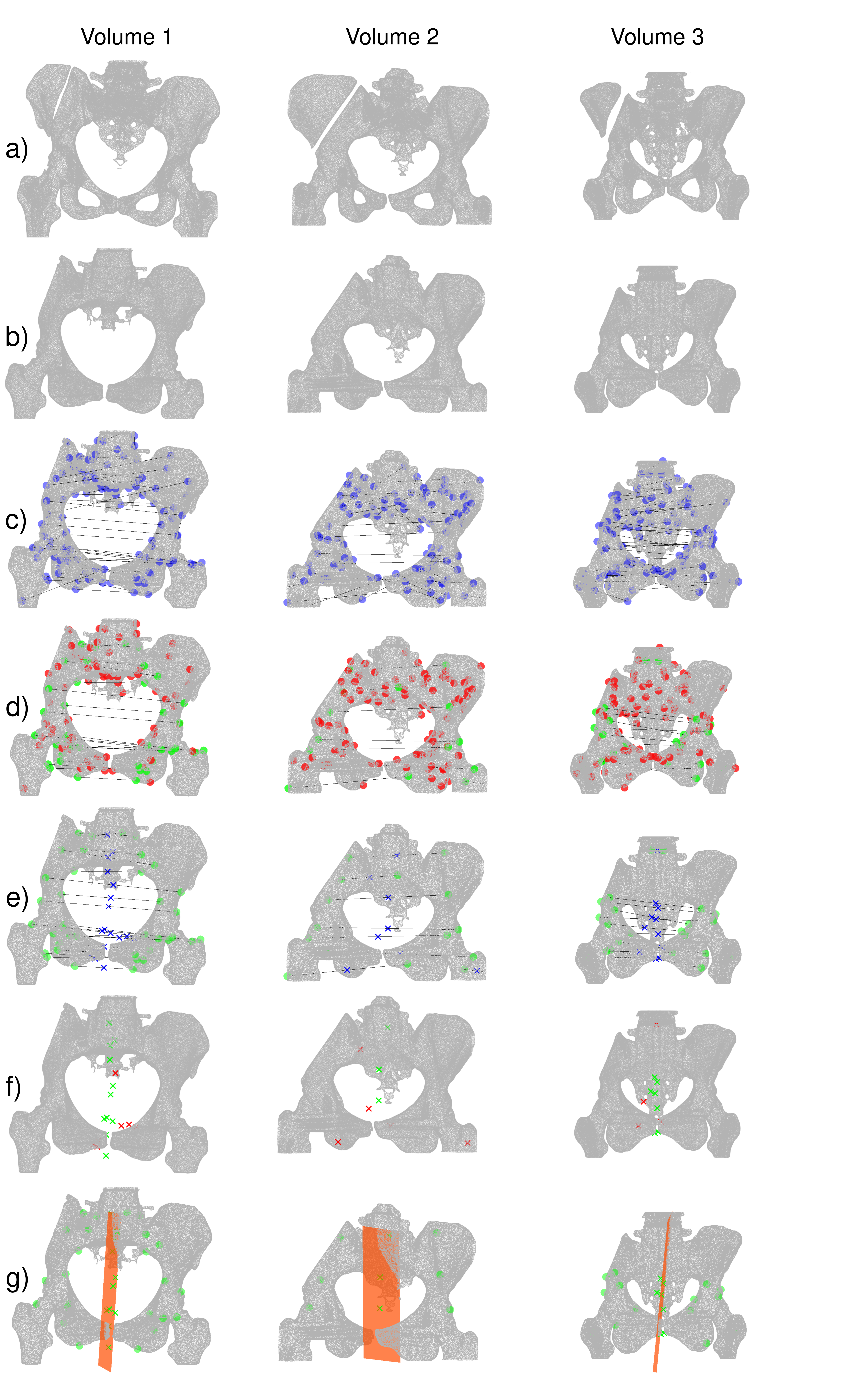}
	\caption{The bone in the CT image was segmented out using an intensity-based thresholding, and converted to volumetric meshes \textbf{(a)}. These meshes were then passed through a hole-closing pipeline which generated genus zero surfaces \textbf{(b)}. If the injury had yielded detached bone fragments with no connections to the main part of the pelvis, then the dislocated bone was automatically rejected and removed using the hole-closing routine. Next, we detected landmark correspondences by computing the optimal anti-M\"obius map. These landmarks satisfied the properties of intrinsic symmetry on the pelvis surface \textbf{(c)}. After the initial landmarks were detected, correspondences were pruned during a two-stage RANSAC. In the first stage, the correspondences that the direction of their connecting vectors were not in agreement with others were removed from the list \textbf{(d)}. Inliers of this step are shown in green, and outliers in red. The remaining landmarks and their midpoints are presented in \textbf{(e)}. In the second RANSAC step, we removed the landmarks which their midpoints did not sit near a plane that passed through the majority of other correspondences \textbf{(f)}. Given the midpoints of all inliers, a least squares estimate yielded the bilateral symmetry plane \textbf{(d)}.}
	\label{fig:mobius}
\end{figure}

The initialization strategy that leverages the combined properties of intrinsic and extrinsic symmetry (Sec.~\ref{subsec:Mobius}) was investigated to assess whether global initialization parameters yield outcome within the capture range. The errors in plane detection, as well as the statistics for the two RANSAC steps are presented in Table~\ref{table:mobius}. In Fig.~\ref{fig:mobius}, we showcase the outcome of each step for three exemplary pelvis data.


\begin{table}
\centering
\caption{Errors in detecting the bilateral symmetry plane are estimated given the initialization parametrization. The last three columns represent the total number of landmarks, inliers with agreement on the direction of the vectors connecting the correspondence, and inliers with consensus on the mid-points.}
\label{table:mobius}
\resizebox{\columnwidth}{!}{
\begin{tabular}{c|c|c|c|c|c}
\cline{2-6}
\multirow{2}{*}{\textbf{}} & \multirow{2}{*}{\textbf{\begin{tabular}[c]{@{}c@{}}Rotation\\ Error (mm)\end{tabular}}} & \multirow{2}{*}{\textbf{\begin{tabular}[c]{@{}c@{}}Translation\\ Error ($^\circ$)\end{tabular}}} & \multicolumn{3}{c}{\textbf{\# Landmarks}} \\ \cline{4-6} 
 &  &  & \textbf{Initial} & \textbf{RANSAC 1 inliers} & \textbf{RANSAC 2 inliers} \\ \hline
\textbf{Mean} & $7.81$ & $4.09$ & $97.3$ & $22.6$ & $14.6$ \\
\textbf{STD} & $2.13$ & $2.61$ & $3.86$ & $6.54$ & $7.32$ \\ \hline
\end{tabular}
}
\end{table}

\subsection{Estimation of Partial Symmetry on Data with Synthetic Fractures}\label{subsec:5}

\begin{figure}
	\centering
		\includegraphics[width=\columnwidth]{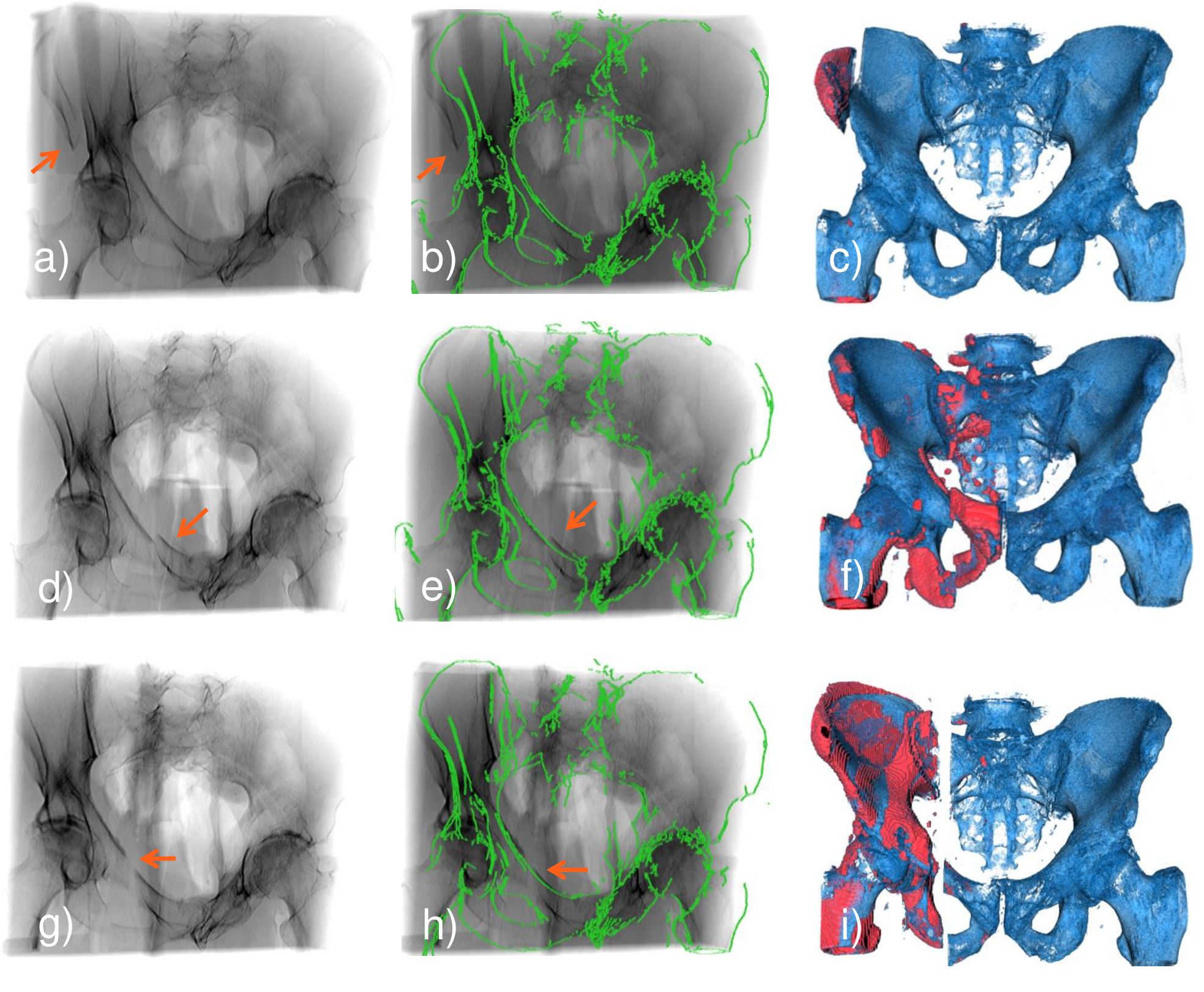}
	\caption{Ilic wing fracture \textbf{(a-c)}, pelvic ring fracture \textbf{(d-f)}, and vertical shear fracture  \textbf{(g-i)} are shown on a pelvis data. The orange arrows in the first two columns represent the area with the fracture. The green contours that are computed from the symmetrically reconstructed model suggest road-maps in each image perspective that can result in fracture reduction and symmetry restoration. The region colored in red in the last column represents the area that was considered as symmetry violator (outlier) by the Tukey-based term $d_I(.)$.}
	\label{fig:synthetic}
\end{figure}

\begin{figure*}[!ht]
	\centering
		\includegraphics[width=\textwidth]{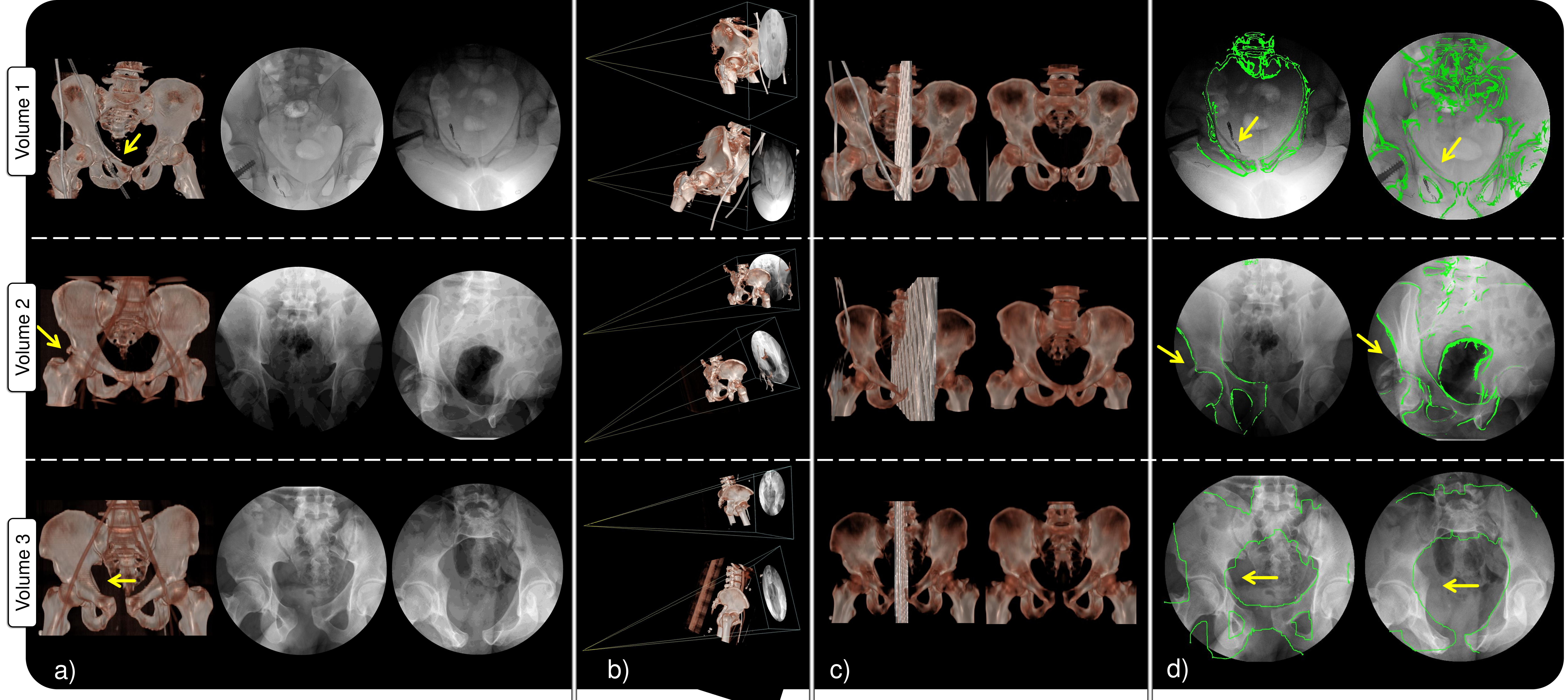}
	\caption{Partial symmetry was exploited to augmented X-ray images with desired configurations of bone fragments. Three cases with severe trauma injuries were investigated here. The first case had displaced pelvic ring fracture, the second case had displaced fracture of the right acetabular roof and the posterior wall, and the third case exhibited asymmetric widening of the left sacroiliac joint. The yellow arrows indicate the fracture location in the figure. We present results given the pre-operative CT as well as two X-ray images \textbf{(a)} for each patient. The relation of each X-ray image was estimated with respect to patient CT \textbf{(b)}. As shown in \textbf{(c)}, the plane of symmetry was detected and symmetric patient-specific template was generated for each case. Finally, DRRs were generated from the symmetric template and were augmented onto patient X-ray as bone outlines.}
	\label{fig:patient}
\end{figure*}

Three prevalent unilateral pelvic fractures, namely, iliac wing fracture, pelvic ring fracture, and vertical shear fracture were applied to four patient data. Given each case, the plane of partial symmetry were detected, and symmetric patient templates were constructed. In Table~\ref{table:landmarkDistance} we report the distance from the anatomical landmark on the original volume before applying the fractures to their reconstructed correspondence on the mirrored model. For the measurements, we considered four separate landmarks that were distributed on the surface of the bone, including, \textbf{L1}: anterior superior iliac spine, \textbf{L2}: posterior superior iliac spine, \textbf{L3}: ischial spine, and \textbf{L4}: ischial ramus. Fig.~\ref{fig:synthetic} shows three simulated radiographs from each fracture model that are augmented with the edge-map extracted from gradient-weighted DRRs of the mirrored template. These contours represent the bone at desired configuration \textit{"if the bilateral symmetry was completely restored"}. This figure also highlights the abnormal area on the bone that was automatically detected using the Tukey-based cost as the symmetry violator region.

\begin{table}
\centering
\caption{Distances between the four anatomical landmarks on the surface of the pelvis with their counterparts on the mirrored template are presented as mean $\pm$ SD.}
\label{table:landmarkDistance}
\resizebox{\columnwidth}{!}{%
\begin{tabular}{l|c|c|c|c}
\hline
\textbf{Fracture} & \textbf{L1 (mm)} & \textbf{L2 (mm)} & \textbf{L3 (mm)} & \textbf{L4 (mm)} \\ \hline
Iliac wing & $3.49 \pm 2.41$ & $2.83 \pm 2.55$ & $3.44 \pm 1.52$ & $2.58 \pm 1.28$ \\
Pelvic ring & $2.24 \pm 1.13$ & $3.96 \pm 2.62$ & $1.78 \pm 0.79$ & $1.84 \pm 0.86$ \\
Vertical shear & $4.96 \pm 1.94$ & $3.35 \pm 2.18$ & $4.07 \pm 2.17$ & $4.52 \pm 2.81$ \\ \hline
\end{tabular}%
}
\end{table}

\subsection{Estimation of Partial Symmetry on Patient Data with Trauma Injuries}\label{subsec:6}

Qualitative outcomes are visualized in Fig.~\ref{fig:patient} for three patient cases with severe unilateral traumatic injuries. 
To the best of our knowledge, no public data set is available with matching CT and X-ray data from unilateral fractures. Therefore, access to larger number of realistic patient data was not possible in this work.
For each case, two fluoroscopic images were separately registered to the patient CT. As the result of 2D/3D registration, the relative projective transformations describing the spatial relation between the X-ray and CT were computed. Next, for all three models, symmetry was detected and patient-specific templates were reconstructed by mirroring the healthy side of the bone across the extrinsic symmetry plane. Finally, DRRs were generated from the patient templates using the projective transformations associate with each X-ray image, and were augmented onto their corresponding fluorscopic images.

\section{Discussion and Conclusion}
In this work we present the methodology for automatic identification of global symmetry in pelvis data with severe unilateral fractures, and exploit the knowledge from symmetry to provide interventional image augmentation. Three measures are combined to identify partial symmetry: \textit{i)} the structural geometry is used in the M\"obius space to determine intrinsic and extrinsic symmetry on the surface, \textit{ii)} voxels are used with Tukey robust estimator to score the similarities between the intensities, and \textit{iii)} normalized mutual information is used to match the distribution of bone across the sagittal plane of the patient. Regularization is important when the amount of bone dislocation is large, and Tukey’s cost cannot solely drive the symmetry plane to the optimal pose. Each of these three novel steps are designed with the consideration of being insensitive to outlier regions that are caused by the injury.

The clinical relevance of this solution is manifested by considering common practices in surgical routine where orthopedic traumatologists aim at bringing displaced bone fragments into alignment with their natural biological configurations. This is achieved by replicating the contralateral side, hence restoring symmetry in the internal structures. It should be noted that this solution is merely admissible for unilateral fractures, that according to pelvis fracture classification~\cite{tile1996acute}, involves a considerable number of cases. Consequently, direct comparison of bony structures across the sagittal plane becomes possible for such cases.

A preeminent criteria in determining symmetry is the Tukey-based robust estimation which automatically suppresses voxel elements that consistently produce high errors. To improve the estimation of the symmetry plane, a novel regularization term based on bone density distribution is added to the overall loss function. In Sec.~\ref{subsec:1} we evaluated the accuracy of symmetry estimation with respect to different regulation factors. Results in Fig.~\ref{fig:tukeyweights} indicate substantial improvement when using regularization. However, the results do not vary significantly when different $\lambda$ factors are used. For consistency, we used $\lambda = 0.5$ for all other experiments in Sec.~\ref{sec:results}.

From the results in Fig.~\ref{fig:captureRange} we conclude that the first three initialization ranges yielded average translation error of $<2\,\text{mm}$ and rotation error of $<0.2^{\circ}$. These are associated with initialization parameters within the ranges of $(0\,\text{mm}, \pm15\,\text{mm})$ translation and $(0^{\circ} , \pm15^{\circ})$ rotation near the ground-truth. A comparison between these results and the errors of the proposed automatic initialization approach presented in Table~\ref{table:mobius}, proves that our suggested initialization yields outcome within the capture range of the cost in Eq.~\ref{eq:cost}. As also appears in Table~\ref{table:mobius}, the correspondence-based initialization demonstrates higher performance in predicting the translation parameters compared to rotation.

We simulated severe and unstable unilateral dislocations, and reconstructed a fully symmetric patient template. Comparing the relevant anatomical landmarks on the original and reconstructed pelvis yielded a mean discrepancy of $3.26$\,mm between different bony features. Finally, we also presented view-specific road-maps to guide towards an optimal repair of pelvic fractures on patient data with trauma injuries. It is essential to highlight here that, because of privacy concerns, and also due to lack of documentation of intra-operative imaging data, the access to matching pairs of CT and X-ray images for unilateral fractures is very limited. Therefore, within the scope of this work, it was only possible to showcase our work on three real data. Future clinical studies may consider more clinical data, if available.

An essential characteristic of our solution is the automatic outlier detection that is highlighted in the last column of Fig.~\ref{fig:synthetic}. Our approach towards outlier identification can enable several other applications in different disciplines of radiology and surgery such (e.g. brain and craniofacial procedures), where regions that violate symmetry can be classified, and consequently be used to improve pre-operative planning as well as provide real-time feedback to surgeons on whether biological symmetry is properly restored. Learning-based solutions can also substantially benefit from such outlier detection mechanisms by automatically shifting the focus of the artificial agent to relevant regions with structural anomaly.

In conclusion, we presented a solution that exploits partial symmetry in human anatomy and provides intuitive image augmentation for fracture care procedures. It should be noted that, this solution enables patient-specific data augmentation and guidance, that is unattainable by using statistical shape models~\cite{chintalapani2010statistical}. Constructing atlases for such procedures require a large population of patient pelvis data for different age, sex, race, disease, etc. which are not available. We hope that our theoretical findings and methodology can lead to safer and more reliable surgical care.

\section*{Acknowledgments}
The authors want to thank Wolfgang Wein and his team from ImFusion GmbH, Munich, for the opportunity of using the ImFusion Suite.





\bibliographystyle{model2-names.bst}\biboptions{authoryear}
\bibliography{refs}

\end{document}